\pdfoutput=1

\documentclass[11pt]{article}
\usepackage{nodalida2025}
\usepackage{times}
\usepackage{url}
\usepackage{latexsym}

\usepackage{amsmath} 
\usepackage[T1]{fontenc}
\usepackage[utf8]{inputenc}
\usepackage{microtype}
\usepackage{inconsolata}
\usepackage{graphicx}
\usepackage{subcaption}
\usepackage{arydshln}
\usepackage{wasysym}
\usepackage{verbatimbox} 
\usepackage{tablefootnote} 
\usepackage{hyperref}

\aclfinalcopy 

\title{OCR Error Post-Correction with LLMs in Historical Documents: No Free Lunches}

\author{Jenna Kanerva, Cassandra Ledins, Siiri K{\"a}pyaho, and Filip Ginter \\
TurkuNLP, Department of Computing \\
University of Turku, Finland \\
\{jmnybl, caledi, siakap, figint\}@utu.fi\\
}

\date{}

\begin{document}
\maketitle
\begin{abstract}
Optical Character Recognition (OCR) systems often introduce errors when transcribing historical documents, leaving room for post-correction to improve text quality. This study evaluates the use of open-weight LLMs for OCR error correction in historical English and Finnish datasets. We explore various strategies, including parameter optimization, quantization, segment length effects, and text continuation methods. Our results demonstrate that while modern LLMs show promise in reducing character error rates (CER) in English, a practically useful performance for Finnish was not reached. Our findings highlight the potential and limitations of LLMs in scaling OCR post-correction for large historical corpora.
\end{abstract}

\section{Introduction}

Digitizing and transcribing historical documents and literature is vital for preserving our cultural heritage and making it accessible for modern digital research methods. The transcription process relies on OCR, which naturally introduces noise into the output. The noise varies in severity depending on the quality of the source material and the OCR technology used, impacting the research usage of the data \citep{chiron2017impact}. The OCR output at two noise levels is illustrated in Figure~\ref{fig:noise-example}. Although modern OCR systems are becoming increasingly accurate \citep{li2022trocr}, reprocessing large collections of historical literature remains a significant challenge, as the resources available to the institutions maintaining these collections are often insufficient for such an undertaking. Consequently, OCR error post-correction has been suggested as means of improving the historical collections without the need to repeat the whole transcription process \citep{nguyen2021survey}.

Recent studies \cite{boros2024postcorrection,bourne2024clocrc} have proposed the application of LLMs to the task, with varying degrees of success. Currently, there is no clear consensus as to how LLMs can be applied to the task and how to deal with the various methodological challenges it poses. Our objective is to address some of these challenges as well as to assess several open LLMs to correct OCR-generated text when prompted to. We study hyperparameter optimization, quantization levels, input lengths, output post-processing and several novel correction methods so as to benchmark and improve the LLM performance on this task.

As our long-term goal is to post-correct two large historical datasets, one in English and the other in Finnish, we focus on these two languages as well as open-weight LLMs, since post-correction of large datasets with commercial models is infeasible cost-wise.

\begin{figure}
\small
    \begin{flushleft}\textbf{Mild noise (0.04 CER):}\end{flushleft}
    \setbox0=\vbox{
    \begin{verbatim}
 A work of art, (be it what it may, house,  
 pi&ure, book, or  garden,) however 
 beautiful in it's underparts, loses half 
 it's value, if the gneralfcope 
 of it be not obvio',s to conception.\end{verbatim} 
    }
    \fbox{\box0}
    \begin{flushleft}\textbf{Severe noise (0.19 CER):}\end{flushleft}
    \setbox1=\vbox{
    \begin{verbatim}
 bke up at Sx in the Mo.r aig.  ll the 
 eauing Withr he went from Cbaud to Cbhh 
 every Suday, «d from Play. bote~PIOoaB 
 cu evi Niuht m the Week, but  vd \end{verbatim} 
    }
    \fbox{\box1}
    \caption{Example extracts of texts at two different OCR noise levels from the ECCO dataset of 18th century literature.}
    \label{fig:noise-example}
\end{figure}


\section{Related work}

Despite decades of active research, post-correction of historical documents remains a challenge. The ICDAR 2017 and 2019 shared tasks \citep{icdar2017,icdar2019} addressed the lack of adequate benchmarks for evaluating OCR performance across several languages, introducing two tracks: detecting OCR errors, and correcting previously detected errors. This setting has naturally guided the development towards two-stage systems, and the best performing models in the ICDAR 2019 edition were based on the BERT model fine-tuned separately for each task \citep{icdar2019}. Such two-step approaches are still actively pursued, with e.g.\ \citet{beshirov2024postocrtextcorrectionbulgarian} applying a BERT classifier for error detection, and an LSTM-based seq2seq model for error correction in Bulgarian.




Recently, LLMs have been effectively applied to text correction problems, for example, \citet{penteado2023evaluating} and \citet{ostling-etal-2024-evaluation} demonstrated that LLMs perform well in grammatical error correction. Naturally, LLMs have been proposed also to the OCR post-correction task, in line with the two broad paradigms of LLM use: fine-tuning for the post-correction task and purely prompt-based zero-shot generation. Fine-tuning was applied e.g.\ by \citet{soper-etal-2021-bart} who fine-tune the BART model on the English subset of the ICDAR 2017 data and apply it to English Newspaper text. \citet{veninga2024msc} fine-tunes the character-based ByT5 model on the ICDAR 2019 data, with a prompt-based Llama model as a baseline. Similarly, \citet{madarasz2024ocrcleaning} apply the mT5 model to historical Hungarian scientific literature, and \citet{dereza-etal-2024-million} applies the BART model to historical Irish--English bilingual data.

In the zero-shot, prompt-based line of work, \citet{boros2024postcorrection} evaluated a variety of models and prompts on several multilingual historical datasets. Interestingly, the results of the study were mostly negative, concluding that LLMs (including the commercial GPT-4 model) are not effective at correcting transcriptions of historical documents, in many cases the LLM actually decreasing the quality instead of improving it. \citet{bourne2024clocrc} conducted a similar study on three historical English datasets, arriving at the opposite conclusion. They achieved over 60\% reduction of character error rate at best, with most of the evaluated models improving the data quality.

Finally, several studies also pursue approaches that include the original image as an input, together with the OCR output to be post-corrected. Here, e.g.\ \citet{chen2024trocrmeets} combine a state-of-the-art transformer-based OCR system with the character-based CharBERT model for handwritten text recognition, and \citet{fahandari2024farsi} propose a model iterating between OCR and post-correction steps for Farsi. Such image-text approaches are, nevertheless, beyond the scope of the present study.





\section{Data}

In our study, we utilize manually corrected samples of two large historical datasets, one for English and the other for Finnish.

\begin{table*}[] 
    \centering
    \begin{tabular}{l|l|rrrrrr}
       Dataset  & Language & Pages & OCR words & GT words & OCR w./pg. & CER & WER  \\\hline
        ECCO-TCP & English & 301,937 & 67,549,822  & 64,519,266 & 223.72 & 0.07 & 0.22 \\
        NLF GT   & Finnish & 449     & 449,088     & 461,305    & 1000.20 & 0.09 & 0.28 \\
    \end{tabular}
    \caption{Dataset statistics after preprocessing in terms of whitespace delimited words. \emph{OCR w./pg.} denotes for mean OCR words per page, and CER and WER are average page-level character and word error rates in the data, weighted by the page length. For details about the metrics, see Section~\ref{sec:metrics}.}
    \label{tab:data-statistics}
\end{table*}


\subsection{English ECCO-TCP}
\label{lab:subsection-ecco}

Eighteenth Century Collections Online (ECCO) \citep{gale_ecco} is a dataset of over 180,000 digitized publications (books and pamphlets) originally printed in the 18th century Britain and its overseas colonies, Ireland, as well as the United States. While mainly in English, some texts appear in other languages. The collection was created by the software and education company Gale by scanning and OCRing the publications. ECCO has significantly impacted 18th-century historical research despite its known limitations \citep{gregg_2021,tolonen2021ecco}.


While ECCO contains only OCR engine output, the ECCO-TCP initiative\footnote{\texttt{https://textcreationpartnership.org/tcp-texts/ ecco-tcp-eighteenth-century-collections-online/}} provides highly accurate, human-made text versions for 2,473 publications from the original collection \citep{gregg2022nature}. In this study, we use a dataset from the Helsinki Computational History Group\footnote{\url{https://github.com/COMHIS}}, where clean ECCO-TCP texts are paired with their corresponding ECCO OCR publications, creating an OCR ground truth dataset \citep{hill2019quantifying}. The data is paired on page level, resulting in a dataset of 338K pages.


To prepare data for post-correction evaluation, we applied several filtering steps. First, we excluded 1,436 pages (0.4\%) marked as blank in ECCO-TCP, ECCO OCR, or both. We also removed 5,782 pages (1.7\%) containing fewer than 150 non-whitespace characters in either collection. Further filtering was necessary in cases of substantial mismatch between OCR and GT pages, typically with large chunks of text missing in either OCR or GT, or otherwise an obvious lack of correspondence. A brief manual analysis identified as typical causes (1) very noisy OCR output with a large amount of non-alphanumerical characters, likely from OCR engine transcribing an image; (2) OCR and GT containing approximately the same text, but in different order due to misidentified reading order or column layout; and (3) significant length differences between pages, possibly from errors in automated page alignments, unrecognized regions left out in the OCR process, or omissions in the ECCO-TCP data. 

To identify such instances, we align each OCR and GT page pair on their non-whitespace characters\footnote{Using global string alignment as implemented in the PairwiseAligner module in biopython.} and slide a 100-character window across the alignment. If in any window less than 10\% of characters were aligned as a match, the page was discarded from the dataset. In total, 28,907 (8.6\%) pages were removed by this process. 

Our filtering produced a dataset of 301,937 well-aligned pages (89.3\% of the initial ECCO-TCP pages). While we do not filter by language, nearly all the data is in English, with other languages appearing only very rarely. 




\subsection{Finnish NLF Ground Truth Data} 

For Finnish, we use the National Library of Finland (NLF) OCR ground truth dataset\footnote{\url{http://digi.nationallibrary.fi}} of \citet{kettunen2018creating,kettunen2020ground}, specifically intended for OCR quality evaluation. The data draws from the National Library's collection of digitized newspapers, and consists of 479 pages randomly chosen from 188 different Finnish newspapers and journals published between 1836--1918, all printed in the Fraktur font. 


The ground truth was created by manually correcting the OCR system output with reference to the original scans. The dataset contains texts produced by three different OCR software (ABBYY FineReader 7/8, ABBYY FineReader 11, and Tesseract) along with the ground truth. In this work, we use output produced by ABBYY FineReader 7/8, which is the OCR engine that has been used to digitize the majority of the NLF collection and therefore gives most useful information for a possible future post-correction effort targeting it.

We applied the same filtering procedure as for the ECCO-TCP data, resulting in the removal of 29 pages (6\%), and we further removed one page written in Swedish. The final dataset consists of 449 pages, with 449,088 OCR words, and 461,305 GT words of text. The key statistics for both datasets are shown in Table~\ref{tab:data-statistics}.




\section{Experiments}
\label{sec:basic-experiments}

First, we set out to evaluate the basic performance of different LLMs on the OCR error correction task and establish how the generation and model hyperparameters (e.g.\ sampling parameters and quantization) affect the results. 

The page lengths in our data vary, with the ECCO-TCP pages on average at 200 words, and the Finnish newspaper pages at about 1,000 words. To improve comparability of the results, we split the pages to segments of 200 OCR words for English and 100 OCR words for Finnish, both corresponding to roughly 300 sub-words in OpenAI's GPT-4 tokenization for the language in question. The length of roughly 300 sub-words was established as suitable in our initial experiments, however, we will carry out a more detailed evaluation of segment lengths as a separate experiment in Section~\ref{sec:segment-length}.

Since the Finnish data is originally word-aligned, obtaining these shorter-than-page segments is trivial. For the English data, which is only page-aligned, we utilize the character-level OCR-GT alignments produced during data filtering (described in Section~\ref{lab:subsection-ecco}), allowing us to find corresponding points. In cases where the segment boundary falls within a region of poor alignment, we shift the boundary to the next reliable word (the word starting the next aligned region). Therefore, the exact segment length may vary depending on how well the OCR and GT strings could be aligned.

Given the substantial volume of our data, and the number of LLM runs necessary in our experiments, we randomly sample for each language a development and a test set, each containing 200 examples (i.e.\ segments of about 300 sub-words in length). These constitute 244K+243K GT characters for English, and 162K+165K GT characters for Finnish. The development set is used to set the generation parameters and the test set is used to report all results, unless otherwise stated.




\subsection{Evaluation Metric}
\label{sec:metrics}


As a primary evaluation metric, we use Character Error Rate (CER) defined as the sum of character substitutions, deletions and insertions, divided by the length of the ground truth string. In line with the common practice in OCR post-correction literature, we mainly report relative CER reduction defined for one examples as $\text{CER\%} = {(\text{CER}_{\text{orig}}-\text{CER}_{\text{post}})}/{\text{CER}_{\text{orig}}}\times 100$
where \emph{orig} and \emph{post} refer to before and after correction, respectively. The overall CER\% is calculated as an average of example-wise CER\% weighted by example lengths in terms of OCR character count. The example-wise CER\% values are further clipped not to go below -100\% to prevent extremely large negative scores in cases where most of the text is omitted. The CER\% therefore works on a range between -100\% and 100\%.

Many downstream applications utilizing historical corpora, such as various literature search interfaces, operate at the level of words rather than characters. Therefore, the main results are reported also in terms of Word Error Rate (WER) and its relative improvement (WER\%). This metric is much like CER, but on the level of words.\footnote{We use the HuggingFace \emph{evaluate} package implementation of both CER and WER.}

Finally, we apply few normalization steps before the evaluation. First, all unicode whitespace characters ($\backslash$s+) are normalized into a single whitespace. Secondly, in line with similar works \citep{duong-etal-2021-unsupervised,kettunen-paakkonen-2016-measuring}, we apply two normalization steps to address systematic differences between historical and modern spellings. In the English ECCO-TCP ground truth data, the long-s character \boxed{\text{\longs}} \ appears in places where modern English would use \boxed{\text{s}}. Similarly, in older Finnish historical texts \boxed{\text{w}} is often used where modern Finnish uses \boxed{\text{v}}. These spelling variations do not alter meaning, and we choose to disregard them by applying Unicode NFKC normalization, which handles both canonically equivalent and compatible transformations (including converting \boxed{\text{\longs}} to \boxed{\text{s}}) for both languages. Additionally, for Finnish, we replace all occurrences of \boxed{\text{w}} with \boxed{\text{v}} before evaluation, as modern Finnish does not use \boxed{\text{w}} except in loanwords or proper names, which occur only very rarely, making the difference negligible. 

\subsection{Models and Generation Parameters}

We evaluate several top-tier open-weight models as well as OpenAI's GPT-4o (v. 2024-08-06). The latter is included mostly for comparison, since it would not be cost-wise feasible to apply it to post-correction at a large scale, unlike open-weight models which can be applied on academic super-computing infrastructure. The evaluated open models are: Llama-3-8B-Instruct, Llama-3.1-8B-Instruct, and Llama-3.1-70B-Instruct from Meta \citep{llama3modelcard,llama3.1modelcard}, Mixtral-8x7B-Instruct-v0.1 from MistralAI \citep{jiang2024mixtral}, and Gemma-2-9B-it and Gemma-2-27B-it from Google \citep{gemma_2024}. It is noteworthy that while several of the open models are multilingual, none officially report supporting Finnish.


All models are run on the Ollama framework\footnote{\url{https://ollama.com/}} (v. 0.3.8) for fast inference, using the default 4-bit quantization unless otherwise stated. Other quantization levels are experimented separately, and reported in Section~\ref{sec:quantization}. All parameters of the framework and models are set to default except for the ones explored in Section~\ref{sec:parameters}. Note that we will not repeat the "Instruct" in model names in tables and figures for space considerations.

\subsection{LLM Overgeneration Removal} \label{lab:subsection-llm-overgen-removal}

LLM outputs often include undesired content in addition to the requested output. In most cases, the undesired text appears either before the corrected text (e.g.\ \emph{"Here is your corrected text:"}), or after the corrected text has been provided (e.g.\ hallucinated continuation, or an additional commentary). This was noted also by \citet{boros2024postcorrection}, who applied simple heuristics for removing any unwanted text, such as removal of whitespace, parts of the prompt, and specific phrases commonly appearing in the model's output, together with trimming the predicted text if it exceeded 1.5 times the original. 

Therefore, we base our overgeneration removal on automatically aligning the generated output against the original input on character level, and filtering out leading and trailing texts which do not align well to the input. For this purpose, we utilize character-level local sequence alignment\footnote{Unlike global sequence alignment, its local counterpart does not penalize leading and trailing misalignments. We use the implementation in the \emph{biopython} package, with open-gap-score -1 and extend-gap-score -0.5} of the model's output and the OCR text, and recover the region between the first and the last aligned characters. The alignment is configured to ignore whitespace and the '-' character, to avoid text formatting discrepancies having an impact on the outcome of the alignment.

\subsection{Parameter Optimization}
\label{sec:parameters}

The model generation parameters naturally affect the quality of the output and we therefore optimize the most critical parameters of the open-weight model generation on a held-out development set. As discussed earlier, this set is not used in any subsequent experiments. 

Using the Optuna hyperparameter optimization library \citep{optuna_2019}, we set the temperature, top\_k and top\_p parameters. For each model and each language, we test 100 runs with different parameter combinations. Subsequently, the 10 best runs in terms of CER were selected, creating a range of possible best parameters. These ranges generally overlap across models but not across languages, therefore we pick a set of parameters for each language. The final parameters are chosen as the median value of the 10 best runs of every model. For English, the parameters are temperature 0.26, top\_k 65, and top\_p 0.66. For Finnish, the final parameters are temperature 0.14, top\_k 30, and top\_p 0.60.



\section{Results}


\begin{table}[]
    \centering
    \begin{tabular}{l|rr|rr}
              & \multicolumn{2}{c}{English} & \multicolumn{2}{c}{Finnish}\\
                      & CER & WER & CER & WER \\
        Model         &  \%  & \%  & \%  & \% \\\hline
        Llama-3-8B    & 7.3 & 31.4 & -68.8 & -28.2 \\
        Llama-3.1-8B  & 19.5 & 37.7 & -65.7 & -30.1 \\
        Llama-3.1-70B & 38.7 & 46.3 & -47.0 & -8.9 \\
        Mixtral-8x7B  & -14.9 & 19.1 & -76.5 & -40.5 \\
        Gemma-2-9B    & 28.2 & 38.4 & -24.0 & -4.1 \\
        Gemma-2-27B   & 35.6 & 37.8 & -19.1 & 0.0 \\
        GPT-4o        & 58.1 & 59.1 & 11.9 & 33.5 \\
    \end{tabular}
    \caption{Overall CER and WER relative improvement.}
    \label{tab:main-results}
\end{table}

\begin{figure}
    \centering
    \includegraphics[width=0.95\linewidth]{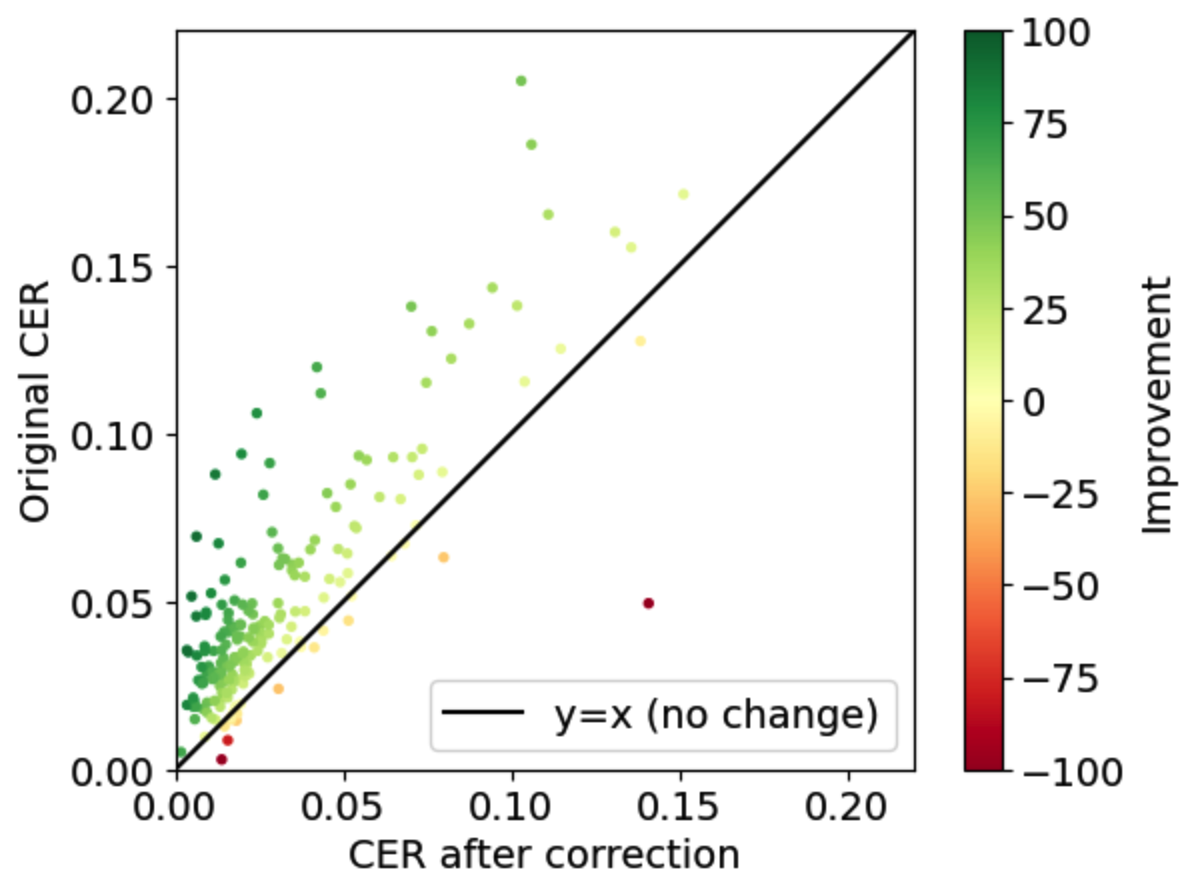}
    \caption{CER before and after correction on English test data (Llama-3.1-70B).}
    \label{fig:before-after-cer}
\end{figure}

The main results are shown in Table~\ref{tab:main-results}. For English, six out of seven models achieve positive improvement, ranging from 7.3\% (Llama-3-8B) to 58.1\% (GPT-4o) in terms of CER\%. GPT-4o outperforms all open models by a large margin, the next best model (Llama-3.1-70B) being almost 20pp worse. However, the Llama-3.1-70B still shows a notable improvement of 38.7\%. In Figure~\ref{fig:before-after-cer} we illustrate the CER values for English test examples before and after the Llama-3.1-70B correction. Most examples demonstrate improved CER scores, regardless of whether they initially had mild or severe noise levels.

In terms of WER\%, all models show positive improvement on English, the two best models achieving an improvement of 59.1\% (GPT-4o) and 46.3\% (Llama-3.1-70B). The relative order of the models seems to mostly follow the number of model parameters, bigger models generally performing better, expect for Mixtral which is clearly worse than the others, and the two Gemma models performing almost equally in terms of WER\%, although the Gemma-2-27B version clearly outperforms the 8B model in terms of CER\%.  

For Finnish, on the other hand, GPT-4o is the only model achieving a positive improvement in either metric, albeit considerably smaller in absolute terms than for English with 11.9 CER\% and 33.5 WER\%. Seeing these entirely negative results for Finnish, we are forced to conclude that prompt-based OCR post-correction is presently infeasible for this language using any of the tested open-weight models. This is disappointing, but not surprising since none of the models officially support Finnish.\footnote{We made also preliminary experiments with the well-known Finnish Poro model of \citet{luukkonen2024poro34bblessingmultilinguality}, but the results were considerably worse than the models in our study, and we did not pursue it any further.}

\begin{figure*}
    \centering
    \includegraphics[width=1\linewidth]{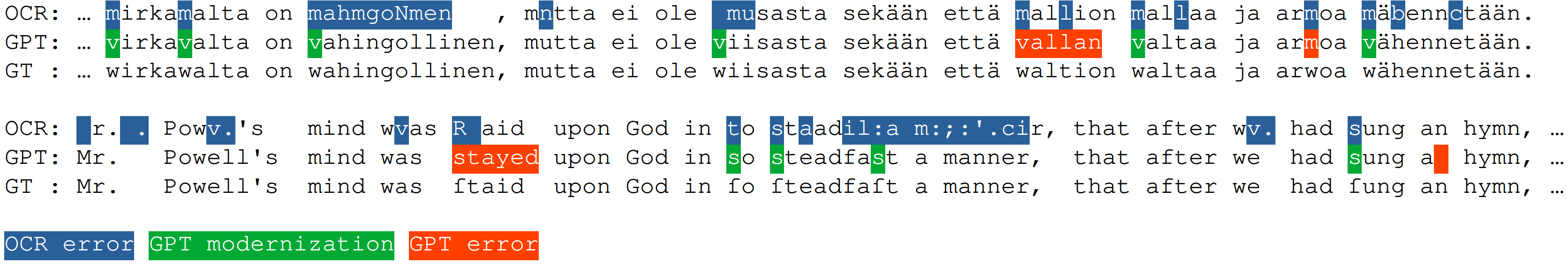}
    \caption{An example in both languages illustrating historical language artifacts alongside the corresponding GPT-4o generated output.}
    \label{fig:example-generation}
\end{figure*}

The striking effect of a common but meaning-preserving difference between historical and modern language becomes apparent when measuring the effect of modern spelling produced by the LLMs such as the \boxed{\text{\longs}} vs.\ \boxed{\text{s}} and \boxed{\text{w}} vs.\ \boxed{\text{v}} variation discussed in Section~\ref{sec:metrics}. A typical example for both languages is illustrated in Figure~\ref{fig:example-generation}. Without the applied normalization, the results of GPT-4o would have been 34.9 CER\% and 35.5 WER\% (compared to 58.1\% and 59.1\% with normalization) for English, and -10.1 CER\% and -4.8 WER\% (compared to 11.9\% and 33.5\%) for Finnish. This demonstrates a substantial impact on the reported scores, and while the relative model ranking is unlikely to change, we can see that the conclusion w.r.t.\ this model's performance on Finnish would have been the opposite, and the improvements seen in English would have been lot smaller.




Given the entirely negative results for Finnish with the open-weight models, we carry out all further analyses on English only. Furthermore, we also remove the Mixtral-8x7B from follow-up experiments as it performs notably worse than the other models.

\subsection{LLM Overgeneration Removal}

\begin{table}[]
    \centering
    \begin{tabular}{l|rr}
    & \multicolumn{2}{c}{Overg. removal}  \\
      Model  & w/o & with  \\
       \hline
        Llama-3-8B & -74.1 & 7.3 \\
        Llama-3.1-8B & -57.4 & 19.5 \\
        Llama-3.1-70B & -53.6 & 38.7 \\
        Gemma-2-9B & 28.1 & 28.2 \\
        Gemma-2-27B & 35.3 & 35.6 \\
        GPT-4o & 53.7 & 58.1 \\
    \end{tabular}
    \caption{The CER improvement on English test data with and w/o the overgeneration removal step.}
    \label{tab:post-processing}
\end{table}



Next we measure the effect of the alignment-based overgeneration removal method described in Section~\ref{lab:subsection-llm-overgen-removal}, i.e.\ we evaluate the raw model-generated output against the post-processed version of the generated output. The results are shown in Table~\ref{tab:post-processing}. For the Llama family models, the results without this post-processing step are highly negative, whereas all Llama models achieve positive improvements when this step is applied. This highlights the necessity of post-processing for the Llama models, which very systematically generate an additional explanation together with the requested output. An example of a typical Llama generation is:

\begin{small}
\begin{verbatim}
 Here is the corrected text: {{answer}}
 I corrected the following errors:
 * "pi\&ure" -> "picture"
 * "it's" -> "its"  (multiple instances)
 * "gneralfcope" -> "general scope"
 ...
\end{verbatim}
\end{small}

On the other hand, Gemma models seem to be largely unaffected, as they generally tend to not produce any additional text. For the GPT-4o model, we also notice a small gain when applying the post-processing, as it occasionally generates explanatory phrases like \emph{"Here is the corrected text:"}.


\subsection{Quantization and Performance}
\label{sec:quantization}

Since the historical text collections to which post-correction would potentially be applied comprise millions of pages of text, it is necessary to strike balance between accuracy and computational resources. Among the most important factors here is model quantization, i.e.\ real number representation with fewer bits. High degrees of quantization substantially reduce model memory footprint and increase inference speed, but can be assumed to potentially degrade model performance. We therefore evaluate the models at the 4 bit Q4\_0 quantization (default setting in Ollama), and at the standard 16 bit fp16 floating point representation. 

The results are reported in Table~\ref{tab:quantization}. As expected, the fp16 quantization performs better for all the evaluated models, with a gain of 2.5-4.7pp, except for Llama-3.1-8B where we do not experience a significant difference between 4bit and fp16 models. The relative ranking of the selected models is preserved regardless the quantization level, and using fp16 does not help less performing models to outrank any of the originally best performing 4bit quantized models. The improvement comes at a high cost in terms of memory footprint. As seen in the table, the best improvement is unsurprisingly achieved by the largest model, where the memory requirement increases from 43Gb to 132Gb. It is of consideration that even with 4bit quantization, using the largest Llama-3.1-70B model would necessitate 2 GPUs (assuming 32GB GPU memory), instantly doubling the GPU hours required to complete the task compared to other models which can fit on one GPU.


        

\begin{table}[]
    \centering
    \begin{tabular}{l|rr|rr}
    & \multicolumn{2}{c|}{CER\%} & \multicolumn{2}{c}{{\centering Memory (Gb)}} \\
        Model & q4 & fp16 & q4 & fp16\phantom{*} \\\hline
        Llama-3-8B & 7.3 & 12.0 & 6.3 & 16.1\phantom{*} \\
        Llama-3.1-8B & 19.5 & 19.4 & 6.3 & 16.1\phantom{*} \\
        Llama-3.1-70B & 38.7 & 42.6 & 43.6 & 132.1* \\
        Gemma-2-9B & 28.2 & 30.7 & 8.9 & 20.9\phantom{*} \\
        Gemma-2-27B & 35.6 & 38.1 & 19.2 & 58.9\phantom{*} \\
    \end{tabular}
    \caption{CER improvement on English test data using 4bit quantized (q4) and fp16 models, alongside peak memory usage. * in the memory consumption indicates the number was obtained using the HuggingFace library, as we were not able to run the Llama-3.1-70B model with fp16 through Ollama.}
    \label{tab:quantization}
\end{table}







\section{Segment Length and Continuation} 
\label{sec:segment-length}

The results in the previous sections were reported for text segments about 300 sub-words in length. The actual texts in the historical collections are naturally substantially longer, necessitating splitting the input into segments of appropriate length. This raises two related questions: (1) how long the input segments should optimally be for best post-correction accuracy, and (2) how should the outputs be combined to minimize degradation on segment boundaries.

Our English data is on the level of pages, which we cannot simply naively concatenate, we need other means to obtain sufficiently long documents. For this experiment, we sample long pages of at least 600 whitespace delimited OCR words in length from the development data, taking at maximum two pages from any one book. This resulted in a sample of 53 development set pages.


\begin{figure} 
    \centering
    \includegraphics[width=.95\linewidth]{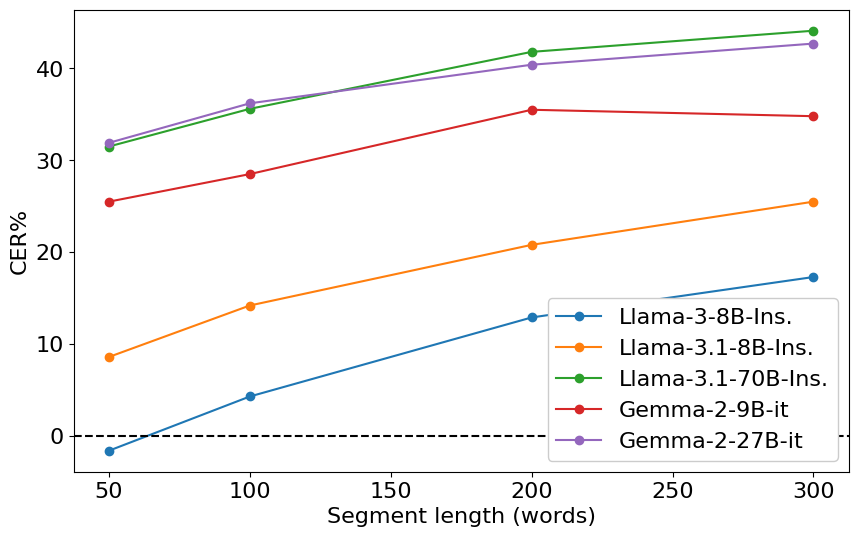}
    \caption{CER\% improvement for English when using different segment lengths.}
    \label{fig:segment-length}
\end{figure}

These sampled pages are then divided into non-overlapping segments of 50, 100, 200, and 300 words, using the same alignment-based splitting strategy as described in Section~\ref{sec:basic-experiments}. The segments are corrected individually and the CER\% improvement over the segments is calculated. The results are shown in Figure~\ref{fig:segment-length}. Shorter segments (50--100 words) get notably worse CER\% score for all models, with the gains diminishing past about 200--300 words, but our page-level data does not have long-enough examples to allow us to reach the point where the performance would start consistently decreasing as the segments become too long. In the future, we plan to develop a book-level version of the data, and study the correction performance on even longer segments.

\subsection{Post-correction on Segment Boundaries}

Presently, post-correction studies either do not address segment-wise correction of longer texts as it is not necessary for the datasets they study, or split the input into non-overlapping segments, whose corrections are simply concatenated. This may potentially disrupt text continuity since neither word nor sentence boundaries can be reliably adhered to in the noisy OCR input. Furthermore, it also means that no left context is available for the correction of the beginning of each segment. This may potentially have a negative effect on the correction in the region around segment boundary. Here we quantify this effect and explore several straightforward methods for its mitigation.


For the prompt optimization, we use the same sample of 53 pages as in the previous section, and the final results are reported on a similar sample of 50 pages from the test data. In order to maximize the number of examples of segment boundaries for evaluation, we generate---from each page---pairs of segments 200+200 words long, with stride of 100 words. With this method, each page of at least 600 words produces a minimum of 3 examples of neighboring segment pairs. The final development and test samples include 194 and 208 such examples, respectively.

On these examples, we evaluate the following methods of post-correction on segment boundaries: (1) {\it Baseline}: Each segment is corrected independently with the same prompt and the outputs are concatenated; (2) {\it Left-corrected-concatenate (LCC)}: The left segment is corrected first and given as prior context for the correction of the right segment, the model is instructed to only correct the right segment; and (3) {\it Left-uncorrected-concatenate (LUC)}: The uncorrected left segment is provided for context in the correction of the right segment. The primary advantage of this method is that correcting the right segment does not need the left segment to be corrected first, making parallelization of the process much simpler technically.

The overall results across these strategies are shown in Table~\ref{tab:merging-overall} and suggest that at present only the two largest models are able to follow the more complex prompts necessitated when merging neighboring segment corrections. The smaller models occasionally suffer from omitting part of the text to correct, which did not seem to occur if only one text was given at a time. For the two models improving on performance, we inspect the boundary effect more closely in Table~\ref{tab:cer-on-seams} where we report CER\% calculated on $\pm 10$ words around the boundary of the two segments.\footnote{The $\pm 10$ word boundaries were human-verified to ensure that the evaluation occurred at the same boundary, even in more complex examples where words were omitted and/or added.} Here we see that both methods effectively incorporate the provided additional context, substantially improving the post-correction of the right segment at the boundary, whereas the baseline system's performance on the right side is notably worse compared to its performance on the left side.


\begin{table}[t]
    \centering
    \begin{tabular}{l|rrr}
              & \multicolumn{3}{c}{Method}\\
        Model & Bas. &  LCC & LUC \\\hline
        Llama-3-8B & -0.2 & -2.9 & -2.8 \\
        Llama-3.1-8B & 13.7 & 8.4 & 10.6 \\
        Llama-3.1-70B & 33.2 & 36.0 & 34.6 \\
        Gemma-2-9B & 29.2 & 27.9 & 28.3 \\
        Gemma-2-27B & 38.1 & 39.9 & 39.7 \\
    \end{tabular}
    \caption{CER improvement on English test data with different correction methods.}
    \label{tab:merging-overall}
\end{table}



\begin{table}[h]
    \centering
    \begin{tabular}{l|cc|c|c}
               & \multicolumn{4}{c}{Method}\\\hline
              & \multicolumn{2}{c|}{Baseline} 
              & LCC 
              & LUC \\
                Model      & L & R   &  R &   R\\\hline
        Llama-3.1-70B & 29.1 &\ 9.8 & 21.4 & 21.7 \\
        gemma-2-27b & 34.5 & 18.7  & 29.7 & 33.7 \\
    \end{tabular}
    \caption{CER\% around segment boundaries with different correction methods. L and R stand for left and right of the boundary.}
    \label{tab:cer-on-seams}
\end{table}



\section{Conclusion}

We set out to establish the ability of recent open-weight models to post-correct OCR errors in a zero-shot, prompt-based setting. In the first set of experiments, we established that for historical English these models achieved notable improvements (Llama-3.1-70B-Instruct reaching a CER improvement of 38.7\%), even though still far behind the commercial GPT-4o model (58.1 CER\%). We also demonstrate the necessity of post-processing to remove any additional, model generated text, and present an effective string alignment technique to address this. We also highlight the effect of segment length, which may have a substantial negative impact on the outcome if set too short.

Unlike for English, for Finnish we find poor performance across the board and need to conclude that zero-shot post-correction with open-weight models remains currently out of reach for historical Finnish. 

In a separate set of experiments we examine how segment-wise correction of long documents should be approached. We devise and evaluate a number of methods to incorporate additional context for the correction of individual segments. We find that some of these methods have a strong positive effect in the immediate proximity of segment boundaries, however, for smaller models the more complicated prompt may cause unexpected degradation in performance when the whole text is considered. Further work will be necessary to resolve these issues.

As future work, we will pursue a large correction run of the ECCO collection as well as a fine-tuned model for Finnish post-correction. All datasets and evaluation scripts used in this study are available at \url{https://github.com/TurkuNLP/ocr-postcorrection-lm} to support result replication and comparability.

\section*{Limitations}

Our work includes certain limitations, which we will discuss next. First, during data preprocessing, we discarded a proportion of documents (\textasciitilde10\% for English, \textasciitilde6\% for Finnish) that our correction methods may not be able to address. These documents include cases with severe alignment issues between OCR output and ground truth. We acknowledge that our post-correction method, which relies entirely on the OCR system's output, cannot recover text where significant portions are missing, therefore setting an upper-boundary for the method. Further analysis is needed to investigate the causes of these gaps and to determine how much, if any, of this missing information could potentially be addressed through post-correction.

We also find that OCR post-correction evaluation suffers from various dataset and metric issues, some of which we have already discussed (e.g.\ normalization). In related work (including our own study conducted directly on our long-term target corpora), results are reported on varying datasets and evaluations metrics. These challenges make it difficult to achieve comparable results across studies and languages, potentially contributing to some of the contradictory conclusions reported in prior work. Clearly, more work will be needed to establish a set of standard benchmarks that resolve most of the data and evaluation issues.

Finally, reporting pure numeric improvements does not address all aspects of downstream data usability. While an improved word error rate has a direct, positive effect on certain applications (e.g.\ lexical search), its impact on others (e.g.\ close reading) may be less straightforward or proportional.

\section*{Acknowledgments}

This work was carried out in the \textit{Human Diversity} University profilation programme (PROFI-7) of the Research Council of Finland, as well as in the context of several other research projects supported by the Research Council of Finland. Computational resources were provided by CSC --- the Finnish IT Center for Science.

\bibliographystyle{acl_natbib}
\bibliography{llm_ocr_main_article}

\begin{thebibliography}{31}
\providecommand{\natexlab}[1]{#1}

\bibitem[{AI@Meta(2024{\natexlab{a}})}]{llama3modelcard}
AI@Meta. 2024{\natexlab{a}}.
\newblock \href {{https://github.com/meta-llama/\\llama3/blob/main/MODEL\_CARD.md}} {Llama 3 model card}.

\bibitem[{AI@Meta(2024{\natexlab{b}})}]{llama3.1modelcard}
AI@Meta. 2024{\natexlab{b}}.
\newblock \href {{https://github.com/meta-llama/llama-models/blob/main/models/llama3\_1/MODEL\_\\CARD.md}} {Llama 3.1 model card}.

\bibitem[{Akiba et~al.(2019)Akiba, Sano, Yanase, Ohta, and Koyama}]{optuna_2019}
Takuya Akiba, Shotaro Sano, Toshihiko Yanase, Takeru Ohta, and Masanori Koyama. 2019.
\newblock Optuna: A next-generation hyperparameter optimization framework.
\newblock In \emph{Proceedings of the 25th {ACM} {SIGKDD} International Conference on Knowledge Discovery and Data Mining}.

\bibitem[{Beshirov et~al.(2024)Beshirov, Dobreva, Dimitrov, Hardalov, Koychev, and Nakov}]{beshirov2024postocrtextcorrectionbulgarian}
Angel Beshirov, Milena Dobreva, Dimitar Dimitrov, Momchil Hardalov, Ivan Koychev, and Preslav Nakov. 2024.
\newblock Post-{OCR} text correction for {Bulgarian} historical documents.
\newblock \emph{ArXiv preprint arXiv:2409.00527}.

\bibitem[{Boros et~al.(2024)Boros, Ehrmann, Romanello, Najem-Meyer, and Kaplan}]{boros2024postcorrection}
Emanuela Boros, Maud Ehrmann, Matteo Romanello, Sven Najem-Meyer, and Frédéric Kaplan. 2024.
\newblock Post-correction of historical text transcripts with large language models: An exploratory study.
\newblock In \emph{Proceedings of the 8th Joint SIGHUM Workshop on Computational Linguistics for Cultural Heritage, Social Sciences, Humanities and Literature (LaTeCH-CLfL 2024)}, pages 133--159. Association for Computational Linguistics.

\bibitem[{Bourne(2024)}]{bourne2024clocrc}
Jonathan Bourne. 2024.
\newblock {CLOCR-C}: Context leveraging {OCR} correction with pre-trained language models.
\newblock \emph{ArXiv preprint arXiv:2408.17428}.

\bibitem[{Chen and Str{\"o}bel(2024)}]{chen2024trocrmeets}
Yung-Hsin Chen and Phillip~B. Str{\"o}bel. 2024.
\newblock {TrOCR} meets language models: An end-to-end post-correction approach.
\newblock In \emph{Proceedings of the Document Analysis and Recognition -- ICDAR 2024 Workshops}, pages 12--26. Springer Nature Switzerland.

\bibitem[{Chiron et~al.(2017{\natexlab{a}})Chiron, Doucet, Coustaty, and Moreux}]{icdar2017}
Guillaume Chiron, Antoine Doucet, Micka{\"e}l Coustaty, and Jean-Philippe Moreux. 2017{\natexlab{a}}.
\newblock {ICDAR} 2017 competition on post-{OCR} text correction.
\newblock In \emph{Proceedings of the 14th IAPR International Conference on Document Analysis and Recognition (ICDAR 2017)}, volume~1, pages 1423--1428. IEEE.

\bibitem[{Chiron et~al.(2017{\natexlab{b}})Chiron, Doucet, Coustaty, Visani, and Moreux}]{chiron2017impact}
Guillaume Chiron, Antoine Doucet, Mickael Coustaty, Muriel Visani, and Jean-Philippe Moreux. 2017{\natexlab{b}}.
\newblock Impact of {OCR} errors on the use of digital libraries: Towards a better access to information.
\newblock In \emph{Proceedings of the 2017 ACM/IEEE Joint Conference on Digital Libraries (JCDL)}, pages 1--4.

\bibitem[{Dereza et~al.(2024)Dereza, N{\'\i}~Chonghaile, and Wolf}]{dereza-etal-2024-million}
Oksana Dereza, Deirdre N{\'\i}~Chonghaile, and Nicholas Wolf. 2024.
\newblock {``}{To} have the {`}million{'} readers yet{''}: Building a digitally enhanced edition of the bilingual {I}rish-{E}nglish newspaper {An Gaodhal} (1881-1898).
\newblock In \emph{Proceedings of the Third Workshop on Language Technologies for Historical and Ancient Languages (LT4HALA) @ LREC-COLING-2024}, pages 65--78. ELRA and ICCL.

\bibitem[{Duong et~al.(2021)Duong, H{\"a}m{\"a}l{\"a}inen, and Hengchen}]{duong-etal-2021-unsupervised}
Quan Duong, Mika H{\"a}m{\"a}l{\"a}inen, and Simon Hengchen. 2021.
\newblock An unsupervised method for {OCR} post-correction and spelling normalisation for {F}innish.
\newblock In \emph{Proceedings of the 23rd Nordic Conference on Computational Linguistics (NoDaLiDa)}, pages 240--248. Link{\"o}ping University Electronic Press.

\bibitem[{Fahandari et~al.(2024)Fahandari, Asadi~Zeydabadi, Shabaninia, and Nezamabadi-pour}]{fahandari2024farsi}
Ali Fahandari, Fatemeh Asadi~Zeydabadi, Elham Shabaninia, and Hossein Nezamabadi-pour. 2024.
\newblock Enhancing {Farsi} text recognition via iteratively using a language model.
\newblock In \emph{Proceedings of the 20th CSI International Symposium on Artificial Intelligence and Signal Processing (AISP)}.

\bibitem[{Gale()}]{gale_ecco}
Gale.
\newblock \href {https://www.gale.com/intl/primary-sources/eighteenth-century-collections-online} {{Eighteenth Century Collections Online}}.

\bibitem[{Gregg(2021)}]{gregg_2021}
Stephen~H. Gregg. 2021.
\newblock \emph{Old Books and Digital Publishing: Eighteenth-Century Collections Online}.
\newblock Elements in Publishing and Book Culture. Cambridge University Press.

\bibitem[{Gregg(2022)}]{gregg2022nature}
Stephen~H. Gregg. 2022.
\newblock The nature of {ECCO-TCP}.
\newblock \emph{Digital Defoe: Studies in Defoe \& His Contemporaries}, 14(1).

\bibitem[{Hill and Hengchen(2019)}]{hill2019quantifying}
Mark~J. Hill and Simon Hengchen. 2019.
\newblock {Quantifying the impact of dirty {OCR} on historical text analysis: Eighteenth Century Collections Online as a case study}.
\newblock \emph{Digital Scholarship in the Humanities}, 34(4):825--843.

\bibitem[{{Jiang} et~al.(2024){Jiang}, {Sablayrolles}, {Roux}, {Mensch}, {Savary}, {Bamford}, {Chaplot}, de~las {Casas}, {Hanna}, {Bressand} et~al.}]{jiang2024mixtral}
Albert~Q. {Jiang}, Alexandre {Sablayrolles}, Antoine {Roux}, Arthur {Mensch}, Blanche {Savary}, Chris {Bamford}, Devendra~Singh {Chaplot}, Diego de~las {Casas}, Emma~Bou {Hanna}, Florian {Bressand}, et~al. 2024.
\newblock Mixtral of experts.
\newblock \emph{arXiv preprint arXiv:2401.04088}.

\bibitem[{Kettunen et~al.(2018)Kettunen, Kervinen, and Koistinen}]{kettunen2018creating}
Kimmo Kettunen, Jukka Kervinen, and Mika Koistinen. 2018.
\newblock Creating and using ground truth {OCR} sample data for {Finnish} historical newspapers and journals.
\newblock In \emph{Proceedings of the Digital Humanities in the Nordic Countries Conference}.

\bibitem[{Kettunen et~al.(2020)Kettunen, Koistinen, and Kervinen}]{kettunen2020ground}
Kimmo Kettunen, Mika Koistinen, and Jukka Kervinen. 2020.
\newblock Ground truth {OCR} sample data of {Finnish} historical newspapers and journals in data improvement validation of a {re-OCRing} process.
\newblock \emph{LIBER Quarterly: The Journal of the Association of European Research Libraries}, 30(1):1--20.

\bibitem[{Kettunen and P{\"a}{\"a}kk{\"o}nen(2016)}]{kettunen-paakkonen-2016-measuring}
Kimmo Kettunen and Tuula P{\"a}{\"a}kk{\"o}nen. 2016.
\newblock Measuring lexical quality of a historical {F}innish newspaper collection ― analysis of garbled {OCR} data with basic language technology tools and means.
\newblock In \emph{Proceedings of the Tenth International Conference on Language Resources and Evaluation ({LREC}'16)}, pages 956--961. European Language Resources Association (ELRA).

\bibitem[{Li et~al.(2023)Li, Lv, Chen, Cui, Lu, Florencio, Zhang, Li, and Wei}]{li2022trocr}
Minghao Li, Tengchao Lv, Jingye Chen, Lei Cui, Yijuan Lu, Dinei Florencio, Cha Zhang, Zhoujun Li, and Furu Wei. 2023.
\newblock {TrOCR}: Transformer-based optical character recognition with pre-trained models.
\newblock In \emph{Proceedings of The Thirty-Seventh AAAI Conference on Artificial Intelligence (AAAI-23)}.

\bibitem[{Luukkonen et~al.(2024)Luukkonen, Burdge, Zosa, Talman, Komulainen, Hatanpää, Sarlin, and Pyysalo}]{luukkonen2024poro34bblessingmultilinguality}
Risto Luukkonen, Jonathan Burdge, Elaine Zosa, Aarne Talman, Ville Komulainen, Väinö Hatanpää, Peter Sarlin, and Sampo Pyysalo. 2024.
\newblock Poro 34b and the blessing of multilinguality.
\newblock \emph{ArXiv preprint arXiv:2404.01856}.

\bibitem[{Madar{\'a}sz et~al.(2024)Madar{\'a}sz, Ligeti-Nagy, Holl, and V{\'a}radi}]{madarasz2024ocrcleaning}
G{\'a}bor Madar{\'a}sz, No{\'e}mi Ligeti-Nagy, Andr{\'a}s Holl, and Tam{\'a}s V{\'a}radi. 2024.
\newblock {OCR} cleaning of scientific texts with {LLMs}.
\newblock In \emph{Natural Scientific Language Processing and Research Knowledge Graphs}, pages 49--58. Springer Nature Switzerland.

\bibitem[{Mesnard et~al.(2024)Mesnard, Hardin, Dadashi, Bhupatiraju, Sifre, Rivière, Kale, Love, Tafti, Hussenot et~al.}]{gemma_2024}
Thomas Mesnard, Cassidy Hardin, Robert Dadashi, Surya Bhupatiraju, Laurent Sifre, Morgane Rivière, Mihir~Sanjay Kale, Juliette Love, Pouya Tafti, Léonard Hussenot, et~al. 2024.
\newblock \href {https://www.kaggle.com/m/3301} {Gemma}.

\bibitem[{Nguyen et~al.(2021)Nguyen, Jatowt, Coustaty, and Doucet}]{nguyen2021survey}
Thi Tuyet~Hai Nguyen, Adam Jatowt, Mickael Coustaty, and Antoine Doucet. 2021.
\newblock Survey of post-{OCR} processing approaches.
\newblock \emph{ACM Comput. Surv.}, 54(6).

\bibitem[{{\"O}stling et~al.(2024){\"O}stling, Gillholm, Kurfal{\i}, Mattson, and Wir{\'e}n}]{ostling-etal-2024-evaluation}
Robert {\"O}stling, Katarina Gillholm, Murathan Kurfal{\i}, Marie Mattson, and Mats Wir{\'e}n. 2024.
\newblock Evaluation of really good grammatical error correction.
\newblock In \emph{Proceedings of the 2024 Joint International Conference on Computational Linguistics, Language Resources and Evaluation (LREC-COLING 2024)}, pages 6582--6593, Torino, Italia. ELRA and ICCL.

\bibitem[{Penteado and Perez(2023)}]{penteado2023evaluating}
Maria~Carolina Penteado and F{\'a}bio Perez. 2023.
\newblock Evaluating {GPT}-3.5 and {GPT}-4 on grammatical error correction for {Brazilian Portuguese}.
\newblock In \emph{Proceedings of the LatinX in AI Workshop at ICML 2023}.

\bibitem[{Rigaud et~al.(2019)Rigaud, Doucet, Coustaty, and Moreux}]{icdar2019}
Christophe Rigaud, Antoine Doucet, Micka{\"e}l Coustaty, and Jean-Philippe Moreux. 2019.
\newblock {ICDAR} 2019 competition on post-{OCR} text correction.
\newblock In \emph{Proceedings of the 15th International Conference on Document Analysis and Recognition (ICDAR 2019)}, pages 1588--1593. IEEE.

\bibitem[{Soper et~al.(2021)Soper, Fujimoto, and Yu}]{soper-etal-2021-bart}
Elizabeth Soper, Stanley Fujimoto, and Yen-Yun Yu. 2021.
\newblock {BART} for post-correction of {OCR} newspaper text.
\newblock In \emph{Proceedings of the Seventh Workshop on Noisy User-generated Text (W-NUT 2021)}, pages 284--290. Association for Computational Linguistics.

\bibitem[{Tolonen et~al.(2021)Tolonen, Mäkelä, Ijaz, and Lahti}]{tolonen2021ecco}
Mikko Tolonen, Eetu Mäkelä, Ali Ijaz, and Leo Lahti. 2021.
\newblock Corpus linguistics and {Eighteenth Century Collections Online} {(ECCO)}.
\newblock \emph{Research in Corpus Linguistics}, 9:19--34.

\bibitem[{Veninga(2024)}]{veninga2024msc}
Martijn Veninga. 2024.
\newblock {LLMs} for {OCR} post-correction.
\newblock Master's thesis, University of Twente.

\end{thebibliography}

\end{document}